\documentclass[runningheads]{llncs}
\usepackage[T1]{fontenc}
\usepackage{graphicx}
\usepackage{cite}
\usepackage{amsmath} 
\usepackage{amsfonts}
\usepackage{tabularx}
\usepackage{url}
\usepackage{booktabs}
\usepackage{multirow}
\usepackage{graphicx}
\usepackage{xcolor}

\begin{document}
\title{HypTI: Hyperbolic Geometry-Aware Multimodal Feature Fusion Framework for Dementia Detection Based on Tabular-Image Data}

%
\titlerunning{HypTI}
%

\author{Jianfei Xu \and
Huizhi Liang \and
Judith Harrison \and
John-Paul Taylor \and
Jaume Bacardit \and
Varun Ojha
}

\authorrunning{Xu et al.}
%
\institute{Newcastle University, Newcastle upon Tyne NE1 7RU, UK \\
\email{\{j.xu65, huizhi.liang, judith.harrison, john-paul.taylor, jaume.bacardit, varun.ojha\}@newcastle.ac.uk}
}
\maketitle              
\begin{abstract}

AI-assisted dementia detection usually requires multimodal fusion of diverse data, such as clinical tabular assessments and neuroimaging data. However, most existing approaches perform feature fusion in Euclidean space. This overlooks the geometric structure of representation, including hierarchical organization in clinical assessments and multi-scale anatomical patterns in neuroimaging data. Recent studies have proved hyperbolic space provides a natural geometry for modelling hierarchical and multi-scale structures due to its exponentially expanding space. Motivated by these, we propose HypTI, a hyperbolic geometry-aware multimodal fusion framework for dementia detection based on clinical tabular assessments and MRI scans. HypTI extends hyperbolic representation learning to an underexplored tabular-MRI setting, jointly modelling clinical tabular, MRI scans, and cross-modal interactions in hyperbolic space. Specifically, clinical tabular variables are represented through learnable Lorentz prototypes to capture hierarchical relationships among assessments, while MRI scans are encoded using ROI-guided multi-scale representation learning to capture anatomical patterns across brain regions and spatial scales in Lorentz space. The multimodal representations are integrated through an adaptive fusion module generates subject-specific scale and shift coefficients to dynamically modulate and integrate the multimodal embeddings, enabling effective patient-level tabular-image representation learning for dementia detection. Experiments on OASIS-3 dataset show that the proposed framework achieves strong performance compared with representative approaches.


\end{abstract}

\section{Introduction}

Dementia syndrome detection requires integrating heterogeneous complementary evidence from clinical assessments and neuroimaging. Clinical tabular variables describe risk factors, cognitive impairments, functional decline, neuropsychiatric symptoms, and clinical staging, whereas T1-weighted MRI provides structural evidence of neurodegeneration and regional brain atrophy\cite{zhang2025b}. Therefore, dementia classification is not only an image recognition problem, but also requires multimodal integration of clinical assessments and neuroanatomical evidence\cite{zhang2011}.

However, learning effective representations from clinical tabular data and MRI remains challenging. The two modalities exhibit substantial differences in their data structure and semantic granularity\cite{baltrusaitis2019}. Clinical tabular variables follow a hierarchical organization consistent with dementia assessment workflows. Broad clinical domains are divided into subdomains and quantified through standardized assessments in a NACC Uniform Data Set (UDS)-style structure, collectively characterizing disease severity\cite{morris2006}. Structural MRI is a high-dimensional 3D signal whose disease-related patterns appear across multiple anatomical scales, ranging from local patch-level tissue changes to region-level atrophy and broader brain-wide structural alterations\cite{pini2016}. In addition, the brain is organized hierarchically across multiple spatial scales\cite{betzel2017}. Voxel-level measurements can be aggregated into anatomical regions, which are in turn embedded within larger structural and functional systems, while neurodegenerative changes often follow non-random spatial patterns across the hierarchy\cite{seeley2009}. Such hierarchical structures are not naturally aligned with flat Euclidean geometry, whose representational capacity grows only polynomially with distance from the origin\cite{nickel2017b}. Most existing multimodal dementia models perform feature fusion in Euclidean space using concatenation-, convolution-, kernel-, or attention-based mechanisms\cite{yang2026,huang2020,odusami2024}. However, such approaches largely overlook the hierarchical and multi-scale geometric structure underlying clinical tabular and neuroimaging representations. Hyperbolic geometry provides a natural alternative for modelling hierarchical and branching structures, as it can compactly organize representations with different levels of specificity within a shared embedding space\cite{nickel2017b,nickel2018a}. Recent advances have extended hyperbolic representation learning to neural networks, graph learning, and vision models\cite{ganea2018,chami2019,ermolov2022}. However, hyperbolic methods have focused on brain connectivity graphs\cite{longhena2025,yang2026}, while joint hyperbolic fusion of clinical tabular data and structural MRI remains largely unexplored.

Motivated by these evidence, we propose a hyperbolic geometry-aware multimodal feature fusion framework for dementia detection based on clinical tabular assessments and MRI scans. Our model treats dementia detection as geometry-aware multimodal disease modelling, where clinical tabular features and multi-scale MRI structural patterns are jointly embedded and fused in Lorentz space. In addition, geometric analyses of the learned Lorentz representations reveal enhanced hierarchical organization, and class separation, providing further insight into the benefits of hyperbolic representation learning.

\textbf{Our~contributions} are summarized as follows: \textbf{(1)} We introduce HypTI, a hyperbolic geometry-aware multi-modal feature fuse framework for medical tabular-image representation learning. \textbf{(2)} We extend medical hyperbolic representation learning to patient-level tabular-MRI modelling, addressing an insufficiently explored multi-modal clinical setting in hyperbolic space. \textbf{(3)} We propose a region-aware multi-scale hyperbolic MRI network (HypMRI-Net) for capturing hierarchical anatomical patterns of brain MRI scans in Lorentz space. \textbf{(4)} We design a hyperbolic geometry-aware fusion module that refines multi-scale MRI embedding through hyperbolic self-attention and generates subject-specific scaling and shifting coefficients to modulate MRI and tabular representations.

\section{Related Work}

\textbf{Multimodal learning for dementia.} Dementia detection is supported by multimodal information, since neurodegeneration is reflected through complementary structural, cognitive, and clinical signals. While single-modality approaches based on MRI, PET, or cognitive assessments have achieved promising results, they provide only a partial view of the disease\cite{frisoni2010, zhang2011, lu2018}. Recent surveys have reviewed multimodal learning for dementia, indicating that that combining neuroimaging, cognitive assessments, and clinical measurements can improve dementia detection\cite{guan2026, catino2026}. However, multimodal dementia studies have shown that conventional feature concatenation and fixed latent aggregation often fail to fully exploit complementary information across modalities\cite{viswan2026}. These limitations motivate more expressive representation learning and feature fusion frameworks for multimodal data, capable of jointly modelling cross-modal dependencies and disease-related structural patterns.

\noindent\textbf{Fusion of tabular and imaging Data.} Existing tabular-imaging fusion methods are categorized into early, late, and intermediate fusion. Early fusion directly concatenates clinical variables with image-derived features, but it often ignores modality-specific structure and allows high-dimensional imaging embeddings to dominate clinical signals\cite{mohsen2022}. Late fusion trains modality-specific predictors and combines their outputs at the decision level, which can improve robustness, but limits cross-modal interaction to the prediction stage. Intermediate fusion combines modality-specific representations in a latent space, but many approaches still rely on concatenation, gating, or coarse conditioning, limiting fine-grained cross-modal interactions \cite{duenias2025}. Recent adaptive fusion strategies, inspired by feature-wise affine modulation\cite{turkoglu2022}, use learned scaling and shifting coefficients to recalibrate multimodal representations and dynamically balance their modality contributions\cite{zhang2025b}. However, these approaches are formulated in Euclidean space and struggle to naturally model the hierarchical structures underlying clinical tabular and multi-scale MRI representations.

\noindent\textbf{Geometry-aware representation learning.} Many conventional representation learning models operate in Euclidean space\cite{yang2026,huang2020,odusami2024}, whose polynomial volume growth limits low-dimensional representations of exponentially branching hierarchies\cite{nickel2017b}. In contrast, hyperbolic space exhibits exponential volume growth with increasing radius, closely matching the geometry of hierarchies and enabling compact representations of complex branching structures \cite{nickel2017b,ganea2018}. Recent dementia studies provide evidence that hyperbolic geometry is useful for representing neuroimaging with hierarchical structure\cite{longhena2025,yang2026}. Longhena et al.\cite{longhena2025} demonstrated that hyperbolic embeddings of DWI and fMRI brain networks can better identify regions disrupted by neurodegeneration than conventional graph metrics. Similarly, hyperbolic graph learning has been explored for neurocognitive decline prediction, where DTI and fMRI brain networks are embedded in hyperbolic space to preserve hierarchical connectivity structures\cite{yang2026}. Despite these advances, existing hyperbolic studies have focused on single image modality, while the joint hyperbolic representation learning of patient-level tabular and image representations remains underexplored. It remains unclear how clinical variables and imaging features can be jointly encoded and fused in hyperbolic space for dementia detection. This motivates a hyperbolic fusion framework that captures both fine-grained tabular–MRI interactions and the hierarchical organization of the resulting multimodal representations.


Overall, Euclidean multimodal models are limited in representing hierarchical and multi-scale structure. Hyperbolic space provides a better geometric match via exponential volume growth and compact low-distortion embeddings of tree-like data\cite{nickel2017b,sala2018}, and hyperbolic attention enables MRI and tabular embeddings to be fused into a flat Euclidean latent space\cite{gulcehre2018,chami2019}. Existing hyperbolic medical models mainly focus on single-modality representations\cite{longhena2025,yang2026}, leaving hyperbolic tabular-MRI fusion underexplored.

\section{HypTI: Hyperbolic Tabular-Image Framework}

\begin{figure}[h]
    \centering
    \includegraphics[width=1\linewidth]{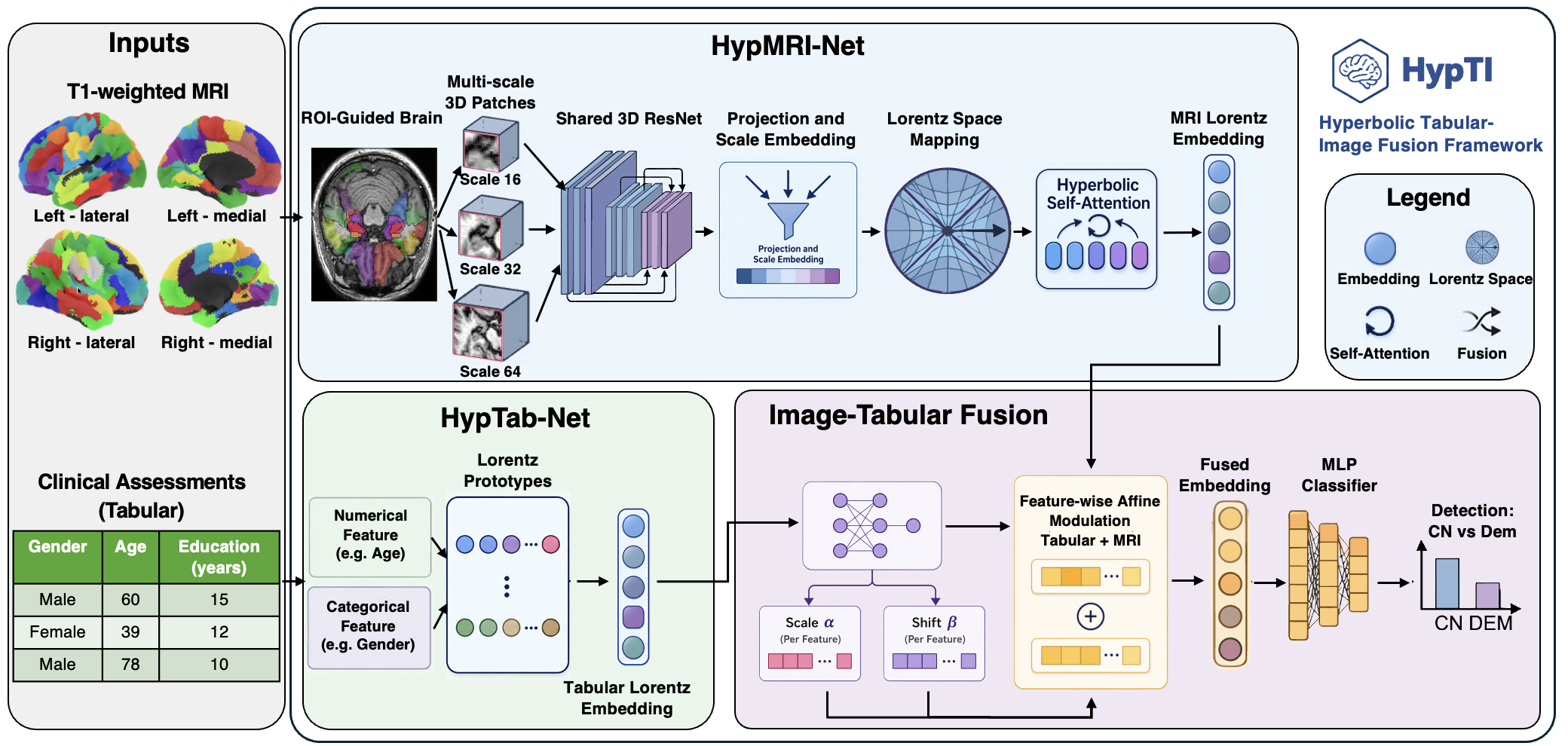}
    \caption{Overview of the proposed HypTI framework.}
    \label{fig:model-architecture}
\end{figure}

The proposed framework (see Figure~\ref{fig:model-architecture}) includes three key modules: First, the Multi-Scale Hyperbolic MRI Network (HypMRI-Net) constructs ROI-guided multi-scale MRI inputs, which are encoded by a 3D ResNet and enhanced via hyperbolic self-attention to model cross-scale anatomical dependencies. Second, the Hyperbolic Tabular Prototype Network (HypTab-Net) represents each tabular variable using a set of learnable Lorentz prototypes and employs a gating mechanism to adaptively aggregate them to form tabular embeddings. Third, an adaptive fusion module generates subject-specific scale and shift coefficients to dynamically integrate multimodal embeddings. The fused tabular–MRI representation is finally fed into an MLP classifier for dementia classification.

\subsection{Problem Formulation}

For each subject, let $X\in\mathbb{R}^{1\times H\times W\times D}$ denote a single-channel T1-weighted MRI volume, where \(H\), \(W\), and \(D\) are the numbers of voxels along the three spatial dimensions. Let $\mathbf{x}\in\mathbb{R}^{F}$ a vector of $F$ clinical variables, $y\in\{0,1\}$ the diagnostic label, and $\mathbf{m}\in\{0,1\}^{F}$ the corresponding availability mask. Here, $m_j=1$ indicates that the $j$-th clinical variable is observed, whereas $m_j=0$ indicates that it is missing. Our objective is to learn a multimodal classifier parameterized by \(\theta\),
$
    \hat{y}=f_{\theta}(X,\mathbf{x},\mathbf{m})
$
, where \(\theta\) denotes the learnable model parameters and \(\hat{y}\) is the predicted diagnostic label.

\subsection{Multi-Scale Hyperbolic MRI Network (HypMRI-Net)}

HypMRI-Net module (see Figure~\ref{fig:model-architecture}) is designed to extract brain anatomical features from MRI patches at multi-scales and capture the relationships among them. Given an input MRI volume $X$, patches are sampled from multiple resolution levels, while an ROI guidance map increases the probability of selecting patches from brain regions highly associated with dementia. Each multi-scale patch is then individually encoded by a shared 3D ResNet backbone to obtain its corresponding patch-level embedding. These patch embeddings are mapped into a common feature space through scale-specific projections and augmented with learnable scale embeddings to indicate their scale type. The scale-aware patch embeddings are subsequently mapped onto the Lorentz manifold to introduce a hierarchical organization of multi-scale MRI embeddings. Finally, hyperbolic self-attention performs matching and aggregation among the patch embeddings, capturing relationships across different brain regions and spatial scales.

\noindent\textbf{ROI-guided multi-scale patch extraction.} For each scale \(s\in\mathcal{S}=\{16,32,64\}\), we extract \(P\) local volumetric patches \(X_{s,p}\) of size \(s\times s\times s\), \(p=1,\ldots,P\), from MRI volume \(X\), yielding \(N_m=|\mathcal{S}|P=3P\) patches in total. During training, patch centres are sampled according to an ROI weight map that assigns higher probabilities to dementia-related regions, including the hippocampus, entorhinal cortex, amygdala, posterior cingulate cortex, precuneus, temporal gyri, and parahippocampal gyrus\footnote{These regions are frequently implicated in Alzheimer's disease(AD) but are not AD-specific and may also be affected in other dementia subtypes. Therefore, for broad dementia detection, the ROI map is treated as an AD-biased anatomical prior that may not fully capture subtype-specific non-AD patterns.}. This sampling strategy exposes the network to local anatomical patterns at complementary spatial resolutions.

\noindent\textbf{Shared 3D ResNet patch encoder.} All patches are encoded by a shared 3D ResNet with three residual blocks. Sharing the backbone across scales encourages a common anatomical feature space while avoiding separate encoders for different patch sizes. A scale-specific learnable projection matrix \(W_s\) maps the encoded features into a common feature space, while a learnable scale embedding \(\mathbf{e}_s\) identifies the spatial resolution of each patch. This yields $\mathbf{u}_{s,p}\in\mathbb{R}^{d}$, the $d$-dimensional scale-aware embedding of \(p\)-th MRI patch sampled at scale \(s\). The resulting patch embeddings are then stacked into \(U_{\mathrm{MRI}}\in\mathbb{R}^{N_m\times d}\):
\begin{equation}
    \mathbf{u}_{s,p}
    =W_s f_{\mathrm{3DResNet}}(X_{s,p})+\mathbf{e}_s,
    \mathbf{u}_{s,p}.
    \label{eq:mri_euclidean_token}
\end{equation}

\noindent\textbf{Lorentz projection of MRI patch embeddings.} We embed each scale-aware MRI patch embedding in a Lorentz manifold of constant negative curvature $-c$, where $c>0$ is the curvature magnitude\cite{ganea2018,ermolov2022}. The $d$-dimensional Lorentz manifold $\mathbb{L}_{c}^{d}$ is:
\begin{equation}
    \mathbb{L}_{c}^{d}
    =\left\{
    \mathbf{x}=(x_0,\mathbf{x}_s)\in\mathbb{R}^{d+1}:
    \langle\mathbf{x},\mathbf{y}\rangle_{\mathcal L}=-\frac{1}{c},\;x_0>0
    \right\},
    \label{eq:lorentz_manifold}
\end{equation}
where $\langle\mathbf{x},\mathbf{y}\rangle_{\mathcal L}$ is the Lorentzian inner product. Subsequently, the point \(\mathbf{o}=(1/\sqrt{c},0,\ldots,0)\) is used as the reference point for mapping Euclidean patch embeddings onto the Lorentz manifold via the exponential map $
    \langle\mathbf{x},\mathbf{y}\rangle_{\mathcal L}
    =-x_0y_0+\mathbf{x}_s^{\top}\mathbf{y}_s.
    \label{eq:lorentz_inner_product}
$

For an MRI patch embedding $\mathbf{u}_{s,p}$, a linear layer first produces a Euclidean vector $\mathbf{z}_{s,p}=W_L\mathbf{u}_{s,p}+\mathbf{b}_L$, where \(W_L\in\mathbb{R}^{d_h\times d_u}\) and \(\mathbf{b}_L\in\mathbb{R}^{d_h}\) are the learnable projection matrix and bias vector, respectively. We lift it to the tangent space at the origin as $\widetilde{\mathbf{z}}_{s,p}=(0,\mathbf{z}_{s,p})\in\mathcal{T}_{\mathbf{o}}\mathbb{L}_c^d$ \(\widetilde{\mathbf z}_{s,p}=(0,\mathbf z_{s,p})\in\mathcal T_{\mathbf o}\mathbb L_c^d\), where \(\mathcal T_{\mathbf o}\mathbb L_c^d\) denotes the tangent space of the Lorentz manifold at the origin \(\mathbf o\). We then apply the exponential map $\exp_{\mathbf{o}}^{c}(\mathbf{v})$, where $\lVert\mathbf{v}\rVert_{\mathcal L}=\sqrt{\langle\mathbf{v},\mathbf{v}\rangle_{\mathcal L}}$ for a tangent vector:
\begin{equation}
    \exp_{\mathbf{o}}^{c}(\mathbf{v})
    =
    \cosh\!\left(\sqrt{c}\lVert\mathbf{v}\rVert_{\mathcal L}\right)\mathbf{o}
    +
    \frac{\sinh\!\left(\sqrt{c}\lVert\mathbf{v}\rVert_{\mathcal L}\right)}
    {\sqrt{c}\lVert\mathbf{v}\rVert_{\mathcal L}}\mathbf{v}.
    \label{eq:lorentz_expmap}
\end{equation}

The Lorentz MRI point is therefore
$
    \mathbf{x}^{L}_{s,p}
    =\exp_{\mathbf{o}}^{c}\!\left(\widetilde{\mathbf{z}}_{s,p}\right)
    \in\mathbb{L}_{c}^{d}.
    \label{eq:mri_lorentz_projection}
$
For compatibility with standard neural layers, we store only its spatial coordinates,
$\mathbf{m}_{s,p}=[\mathbf{x}^{L}_{s,p}]_{1:d}$. The time-like coordinate can be recovered without loss as:
\begin{equation}
    [\mathbf{x}^{L}_{s,p}]_0
    =\sqrt{\lVert\mathbf{m}_{s,p}\rVert_2^2+\frac{1}{c}}.
    \label{eq:recover_time_coordinate}
\end{equation}
The Lorentz MRI patch embedding sequence $M_{\mathrm{MRI}}$ is denoted by $
M_{\mathrm{MRI}}=\{\mathbf{m}_{s,p}\}_{s\in\mathcal S,p=1}^{P}\in\mathbb{R}^{N_m\times d}.
$

\noindent\textbf{Lorentz-aware MRI self-attention.} Hyperbolic self-attention models dependencies among anatomical patches and scales. From each reconstructed Lorentz patch embedding, learnable projections produce query, key, and value points $\mathbf{q}_i,\mathbf{k}_i,\mathbf{v}_i\in\mathbb{L}_c^d$. Query--key compatibility is measured using the Lorentz geodesic distance $
    d_{\mathbb{L}_c}(\mathbf{x},\mathbf{y})
    =\frac{1}{\sqrt{c}}
    \operatorname{arcosh}\!\left(
    -c\langle\mathbf{x},\mathbf{y}\rangle_{\mathcal L}
    \right),
    \label{eq:lorentz_distance}
$ and the attention weights are
$
    A^{m}_{ij}
    =\operatorname{softmax}_{j}\!\left(
    -d_{\mathbb{L}_c}^{2}(\mathbf{q}_i,\mathbf{k}_j)
    \right).
    \label{eq:hyperbolic_attention_weights}
$

To preserve the geometry during value aggregation, we use the Einstein midpoint rather than a Euclidean weighted sum\cite{chami2019}. A Lorentz point $\mathbf{v}=(v_0,\mathbf{v}_s)$ is first mapped to its Klein coordinate
$\boldsymbol{\kappa}(\mathbf{v})=\mathbf{v}_s/v_0$. The output associated with query $i$ is:
\begin{equation}
    \boldsymbol{\kappa}_i^{\ast}
    =
    \frac{
    \sum_j A^{m}_{ij}\gamma_j\boldsymbol{\kappa}(\mathbf{v}_j)
    }{
    \sum_j A^{m}_{ij}\gamma_j
    },
    \qquad
    \gamma_j=\frac{1}{\sqrt{1-\lVert\boldsymbol{\kappa}(\mathbf{v}_j)\rVert_2^2}},
    \label{eq:einstein_midpoint}
\end{equation}
which is mapped back to the Lorentz model\cite{gulcehre2018}. Collecting the spatial coordinates of the refined points gives
$M^{\ast}=\{\mathbf{m}^{\ast}_i\}_{i=1}^{N_m}$.

\subsection{Hyperbolic Tabular Prototype Network (HypTab-Net)}
\label{subsec:hyptab}

HypTab-Net (see Figure~\ref{fig:model-architecture}) represents each clinical variable independently following the ProtoGate mechanism used in HDE-Net\cite{li2026}, so that its identity and clinical role are retained during fusion. For feature $j$, we assign a learnable prototype bank
$\mathcal{P}_j=\{\mathbf{p}_{j,k}\}_{k=1}^{K}$, where each
$\mathbf{p}_{j,k}\in\mathbb{L}_c^{d_t}$ is constrained to the Lorentz manifold in Equation~\eqref{eq:lorentz_manifold}. For the \(j\)-th clinical variable, a feature-specific learnable gating function \(g_j:\mathbb R\rightarrow\Delta^{K-1}\) converts its scalar value \(x_j\) into mixing coefficients $
    \boldsymbol{\alpha}_j(x_j)
    =g_j(x_j)
    \label{eq:tabular_gate}
$, where \(\alpha_{j,k}\) denotes the weight assigned to the \(k\)-th prototype, with \(\alpha_{j,k}\geq0\) and \(\sum_{k=1}^{K}\alpha_{j,k}=1\).
Because Lorentz points cannot be directly combined using Euclidean addition, the prototypes are first mapped to the tangent space at the origin. For
$a_{j,k}=-c\langle\mathbf{o},\mathbf{p}_{j,k}\rangle_{\mathcal L}$, the logarithmic map is:
\begin{equation}
    \log_{\mathbf{o}}^{c}(\mathbf{p}_{j,k})
    =
    \frac{\operatorname{arcosh}(a_{j,k})}
    {\sqrt{a_{j,k}^{2}-1}}
    \left(\mathbf{p}_{j,k}-a_{j,k}\mathbf{o}\right).
    \label{eq:lorentz_logmap}
\end{equation}
The gated prototype mixture is formed in the tangent space and returned to the manifold using the exponential map in Equation~\eqref{eq:lorentz_expmap}:
\begin{equation}
    \mathbf{h}_j
    =\exp_{\mathbf{o}}^{c}\!\left(
    \sum_{k=1}^{K}
    \alpha_{j,k}(x_j)
    \log_{\mathbf{o}}^{c}(\mathbf{p}_{j,k})
    \right).
    \label{eq:tabular_prototype_mixture}
\end{equation}
We retain the spatial coordinates of $\mathbf{h}_j$ as its feature-wise tabular representation. To handle missing values, each feature is also assigned a learnable missing embedding $\mathbf{r}_j\in\mathbb{R}^{d_t}$, giving $
    \mathbf{t}_j
    =m_j[\mathbf{h}_j]_{1:d_t}+(1-m_j)\mathbf{r}_j.
    \label{eq:masked_tabular_embedding}
$
The complete tabular sequence is
$T=\{\mathbf{t}_j\}_{j=1}^{F}\in\mathbb{R}^{F\times d_t}$.

\subsection{Geometry-Aware Cross-Modal Fusion}
\label{subsec:fusion}

The fusion module (see Figure~\ref{fig:model-architecture}) integrates the refined MRI patch embeddings $M^{\ast}$ with the feature-wise tabular sequence $T$. We summarize the MRI sequence using average pooling (AvgPool) and max pooling (MaxPool), where AvgPool captures the overall multi-scale brain structure and MaxPool highlights the most discriminative ROI patterns across scales. The tabular sequence is flattened(Flatten) to preserve feature-specific clinical information:
\begin{equation}
    \mathbf{z}_m
    =\left[
    \operatorname{AvgPool}(M^{\ast});
    \operatorname{MaxPool}(M^{\ast})
    \right],
    \qquad
    \mathbf{z}_t=\operatorname{Flatten}(T).
    \label{eq:modality_summaries}
\end{equation}
A learnable fusion network $\phi_f$ predicts subject-specific scale ($\boldsymbol{\alpha}$) and shift ($\boldsymbol{\beta}$) parameters from both modalities, where \(\mathbf{z}_m\) denotes the subject-level MRI representation aggregated from the refined multi-scale MRI patch embeddings, and \(\mathbf{z}_t\) denotes the subject-level tabular representation generated by HypTab-Net $
    [\boldsymbol{\alpha},\boldsymbol{\beta}]
    =\phi_f\!\left([\mathbf{z}_m;\mathbf{z}_t]\right)
    \label{eq:fusion_parameters}
$.
The tabular context then modulates the MRI representation through a feature-wise affine transformation $
    \widetilde{\mathbf{z}}_m
    =\boldsymbol{\alpha}\odot\mathbf{z}_m+\boldsymbol{\beta},
    \mathbf{z}=[\widetilde{\mathbf{z}}_m;\mathbf{z}_t]
    \label{eq:scale_shift_fusion}
$, where $\odot$ denotes the Hadamard product, \(\mathbf z_m\in\mathbb R^{d_m}\) is the original MRI representation, \(\mathbf z_t\in\mathbb R^{d_t}\) is the encoded tabular context, and \(\widetilde{\mathbf z}_m\) is the tabular-conditioned MRI representation. This mechanism enables clinical information to recalibrate MRI-derived features on a subject-specific basis instead of treating the two modalities as equally informative under direct concatenation.

The fused representation \(\mathbf{z}\) is passed to an MLP classification head to produce the multimodal prediction. During training, a tabular-only auxiliary classifier is also applied to the tabular representation \(\mathbf{z}_t\), producing a tabular-only prediction. This auxiliary classifier provides direct label supervision to HypTab-Net, as optimizing only the multimodal objective may cause the model to rely predominantly on MRI features and underutilize the tabular modality. The auxiliary objective therefore encourages \(\mathbf{z}_t\) to retain discriminative clinical information and stabilizes the learning of the tabular branch. The overall objective is $
    \mathcal{L}
    =\mathcal{L}_{\mathrm{multi}}
    +\lambda_{\mathrm{tab}}\mathcal{L}_{\mathrm{tab}}
    \label{eq:training_objective}
$, where $\mathcal{L}_{\mathrm{multi}}$ and $\mathcal{L}_{\mathrm{tab}}$ are the cross-entropy losses for the multimodal and auxiliary tabular predictions, respectively, and $\lambda_{\mathrm{tab}}$ controls the auxiliary contribution. Lorentz prototypes and other manifold-constrained parameters are optimized using Riemannian Adam, whereas unconstrained neural-network parameters are optimized using AdamW.

\section{Experiments}

\subsection{Dataset and Metrics}

We train our model using the OASIS-3 dataset\footnote{\url{https://sites.wustl.edu/oasisbrains/home/oasis-3/}}, which is a multimodal neuroimaging and clinical assessments from individuals spanning normal ageing to dementia. The dataset includes retrospective data from 1,378 participants, including 755 cognitively normal adults and 622 individuals at different stages of cognitive decline. OASIS-3 includes 2,842 MR sessions (4,116 raw T1-w MRI), and tabular data such as demographic variables, UDS clinical forms, cognitive assessments, and psychometric measurements. The OASIS-3 cohort used in our experiment was constructed by first deriving subject-level CN and dementia labels from clinical assessments\footnote{Dementia labels were assigned to subjects with either \texttt{DEMENTED = 1} in \texttt{OASIS3\_UDSd1\_diagnoses.csv} or \texttt{CDR} $\geq$ 1 at any visit. This dementia label definition captures broad clinical dementia status and does not distinguish between different dementia aetiologies or subtypes. CN labels were assigned to subjects in the OASIS-3 cognitively stable cohort (\texttt{OASIS3\_UDSb4\_cdr.csv}) with \texttt{CDR = 0} across all visits. Subjects with MCI or ambiguous diagnostic status were excluded.}, then intersecting 1,254 labelled subjects with those having usable T1-weighted MRI scans, yielding 540 subjects (see Table~\ref{tab:oasis3_stats}).

\begin{table}[h]
\centering
\caption{Cohort statistics and demographic characteristics of the OASIS-3 subset used in this study. Continuous variables are reported as mean $\pm$ standard deviation.}
\begin{tabular}{lcc}
\toprule
\textbf{Variable} & \textbf{Cognitively Normal (CN)} & \textbf{Dementia} \\
\midrule
Gender (Male/Female) & 120/191 & 115/114 \\
Age & $65.45 \pm 9.50$ & $73.45 \pm 7.88$ \\
Education(years) & $15.92 \pm 2.50$ & $15.15 \pm 3.04$ \\
\midrule
Total Subjects &  311 & 229  \\
\bottomrule
\end{tabular}
\label{tab:oasis3_stats}
\end{table}

Performance was evaluated on each test split using Accuracy (ACC), Matthews Correlation Coefficient (MCC), and area under the receiver operating characteristic curve (AUC). We primarily report AUC for threshold-independent discrimination and MCC for balanced binary classification under class imbalance \cite{chicco2020}.

\subsection{Preprocessing}

\noindent\textbf{Tabular data.} Each subject was represented by 37 clinical features derived from OASIS-3 assessments, including demographic, APOE genotype\footnote{APOE genotype describes the inherited apolipoprotein E allele combination, where the $\varepsilon4$ allele is a major genetic risk factor for dementia.}, cognitive scores, and medical histories\cite{zhang2025b}. Continuous features were Z-score normalized using training-split statistics. Missing values were zero-filled in the normalized space and encoded using a 37-dimensional availability mask, enabling the model to distinguish missing entries and exploit informative missingness patterns\cite{che2018}.

\noindent\textbf{MRI data.} T1-weighted MRI scans were converted to BIDS format\footnote{\url{https://bids.neuroimaging.io/index.html}} using Clinica's \texttt{oasis3-to-bids} converter\footnote{\url{https://aramislab.paris.inria.fr/clinica/docs/public/dev/Converters/OASIS3TOBIDS/}} and preprocessed with the \texttt{t1-linear} pipeline\footnote{\url{https://aramislab.paris.inria.fr/clinica/docs/public/dev/Pipelines/T1_Linear/}}. Preprocessing included N4 bias correction, affine registration to the MNI-152 space, 1 mm isotropic resampling, skull stripping, and background cropping, yielding volumes of size $169 \times 208 \times 179$\cite{routier2021}. This placed all subjects in a common anatomical space for downstream modelling.

Before network input, MRI volumes were processed with MONAI\footnote{\url{https://github.com/project-monai/monai}}. Voxel intensities were min-max normalized, and each volume was center-cropped to $160 \times 192 \times 160$. During training, MONAI transforms were applied for data augmentation, including random contrast adjustment ($p=0.5$), rotation ($\pm0.05$ radian, $p=0.2$), and zooming ($[0.95,1.05]$, $p=0.2$)\cite{lu2018}. To capture neuroanatomical patterns, we used ROI-guided patch extraction. For each MRI, 16 patches were sampled at each of three scales ($64^3$, $32^3$, and $16^3$). 

\subsection{Training Details}

All experiments were conducted using 5-fold cross-validation. Training followed a two-stage scheme: the tabular branch was first optimized with an auxiliary classification loss, followed by optimization of the MRI-tabular fusion network. Euclidean parameters were optimized with AdamW ($2\times10^{-4}$), and Lorentz manifold parameters with RiemannianAdam ($10^{-6}$). Model hyperparameters were optimized using Optuna\footnote{https://optuna.org} with 100 trials, and the final model was trained using the best validation configuration. 

\section{Results}

\subsection{Performances}

We compare HypTI with representative machine-learning, deep-learning, and multimodal approaches: \textbf{(1) Machine-learning models} \cite{bloch2022}: Logistic Regression, Decision Tree, Random Forest, XGBoost, and SVM were included as tabular-data baselines, covering linear, tree-based, ensemble, boosting, and kernel-based classification methods. \textbf{(2) Lee et al.} \cite{lee2019}: An MRI-based deep-learning model that integrates regional abnormality representations for dementia-related neurodegeneration. \textbf{(3) Qiu et al.} \cite{qiu2022}: A multimodal deep-learning framework combining structural MRI with demographic and clinical information in Euclidean space. \textbf{(4) Zhang et al.(Baseline)} \cite{zhang2025b}: A modality-flexible framework that integrates imaging and tabular features through self-attention transformer in Euclidean space. \textbf{(5) HypTI (Ours)}: Our framework represents tabular and multi-scale MRI features in Lorentz space and performs multimodal fusion.

\begin{table*}[h]
\centering
\small
\setlength{\tabcolsep}{4pt}
\renewcommand{\arraystretch}{0.95}
\caption{Performance comparison on the OASIS-3 dataset for cognitively normal vs dementia classification. Results are reported as mean $\pm$ standard deviation (\%) using 5-fold cross-validation. The $\ast$ denotes the baseline method from Zhang et al.\cite{zhang2025b}, evaluated on our dataset. Bold values indicate the best performance, and ``--'' indicates metrics not reported by the compared state-of-the-art methods.}
\resizebox{\textwidth}{!}{
\begin{tabular}{llllll}

\toprule

Modality & Method & AUC & ACC & MCC \\

\midrule

\multirow{4}{*}{MRI Only}

& Zhang et al.\cite{zhang2025b}$\ast$ & 81.06 $\pm$ 3.25 & 72.21 $\pm$ 6.09 & 43.55 $\pm$ 12.84 \\

& Lee et al.\cite{lee2019} & -- & 71.10 & -- \\


& Ours &  \textbf{84.19} $\pm$ 4.77 & \textbf{75.05} $\pm$ 3.44 & \textbf{58.32} $\pm$ 7.38 \\

\midrule

\multirow{7}{*}{Tabular Only}

& Zhang et al.\cite{zhang2025b}$\ast$ & 85.59 $\pm$ 7.43 & 83.69 $\pm$ 6.47 & 61.02 $\pm$ 13.81 \\

& Logistic Regression\cite{bloch2022} & 82.16 & 78.94 & 46.60 \\

& Decision Tree\cite{bloch2022} & 76.21 & 61.75 & 22.80 \\

& Random Forest\cite{bloch2022} & 75.52 & 68.96 & 29.20 \\

& XGBoost\cite{bloch2022} & 81.28 & 78.05 & 46.90 \\

& SVM\cite{bloch2022} & 82.33 & 80.71 & 47.40 \\

& Ours & \textbf{86.22} $\pm$ 7.52 & \textbf{85.01} $\pm$ 5.49 & \textbf{62.46} $\pm$ 12.00 \\

\midrule

\multirow{4}{*}{Tabular$+$MRI}

& Zhang et al.\cite{zhang2025b}$\ast$ & 81.94 $\pm$ 7.07 & 77.25 $\pm$ 3.48 & 46.27 $\pm$ 9.06 \\

& Qiu et al.\cite{qiu2022} & -- & 80.40 & -- \\


& Ours & \textbf{90.09} $\pm$ 1.58 & \textbf{83.88} $\pm$ 3.15 & \textbf{67.58} $\pm$ 6.45 \\

\bottomrule

\end{tabular}
}
\label{tab:oasis3-comparison-result}
\end{table*}

Table~\ref{tab:oasis3-comparison-result} compares HypTI with representative methods under different modality settings. In the MRI-only setting, our model achieves the best performance across all metrics, with an AUC of 84.19\%, ACC of 75.05\%, and MCC of 58.32\%. In the tabular-only setting, our model also outperforms both machine learning methods and the baseline model, achieving 86.22\% AUC, 85.01\% ACC, and 62.46\% MCC. When combining tabular and MRI data, the proposed HypTI obtains the best overall performance, reaching 90.09\% AUC, 83.88\% ACC, and 67.58\% MCC. These results demonstrate the effectiveness of hyperbolic representation learning for both single-modality encoding and multi-modal fusion.

\begin{figure}[t]
\centering
\includegraphics[width=0.9\linewidth]{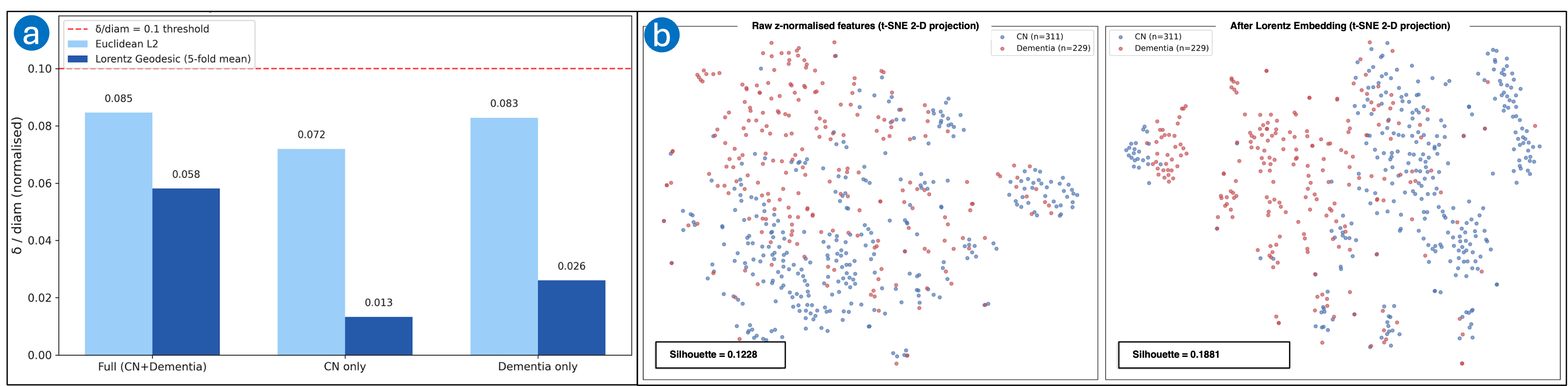}
\caption{Geometric motivation for hyperbolic modelling of clinical tabular data. (a) Gromov $\delta$-hyperbolicity analysis of raw and Lorentz feature spaces. (b) t-SNE visualization before and after Lorentz embedding.}
\label{fig:tabular-tree-structure}
\end{figure}
\begin{figure}[t]
\centering
\includegraphics[width=0.9\linewidth]{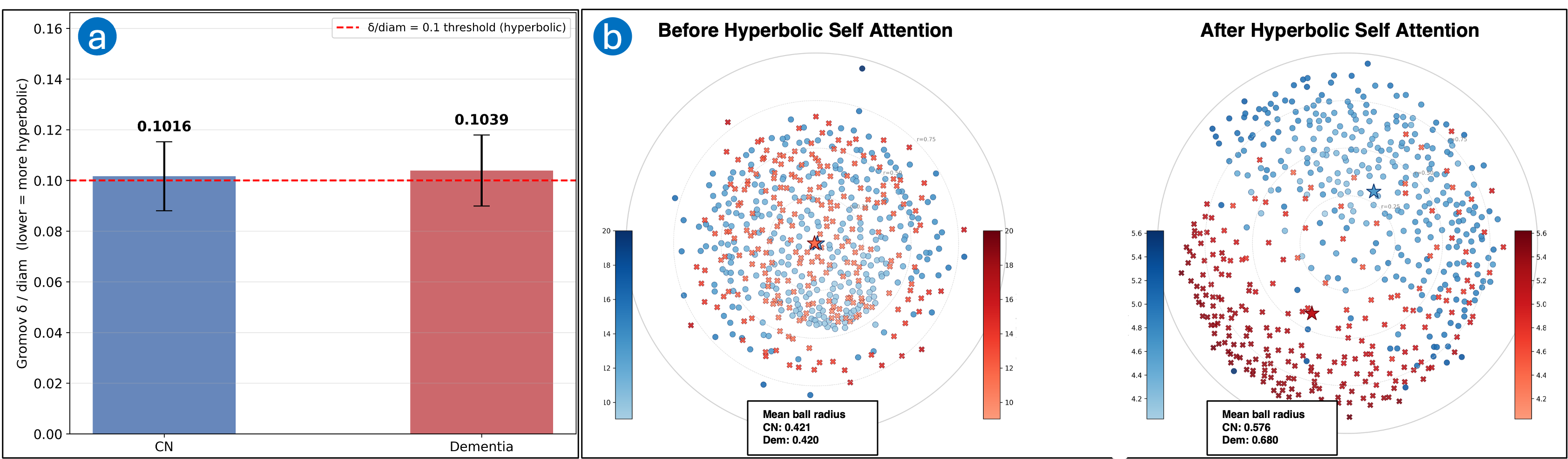}
\caption{Geometric analysis of MRI patch-token representations. (a) Gromov $\delta$-hyperbolicity of MRI token spaces. (b) MDS visualization before and after Lorentz self-attention using Lorentz geodesic distances.}
\label{fig:mri-tree-like}
\end{figure}

To validate the geometric motivation of HypTI, we analyse the intrinsic structure of tabular data and MRI representations using Gromov $\delta$-hyperbolicity~\cite{chalopin2021} and low-dimensional visualizations. For tabular features (Figure~\ref{fig:tabular-tree-structure}), the raw OASIS-3 space already shows tree-like structure with $\delta/\mathrm{diam}<0.1$, and Lorentz embedding further reduces $\delta/\mathrm{diam}$ from 0.085 to 0.058. The t-SNE visualization also shows clearer CN-dementia separation. A similar pattern is observed for MRI representations (Figure~\ref{fig:mri-tree-like}): multi-scale patch tokens have $\delta/\mathrm{diam}\approx0.1$, and Lorentz attention yields clearer class separation in MDS space. These results support the hierarchical geometry of both modalities and show that hyperbolic modelling improves representation separability for dementia classification.

\subsection{Ablation Study}

To validate the contribution of the components, we conducted ablation experiments from three perspectives: Hyperbolic vs Euclidean representation, single-scale vs multi-scale MRI encoding, and ROI-guided vs random patch sampling. 

\begin{table*}[h]
\centering
\small
\setlength{\tabcolsep}{4pt}
\renewcommand{\arraystretch}{0.95}
\caption{Ablation study of Euclidean and hyperbolic representations on OASIS-3 dataset. Results are reported as mean $\pm$ standard deviation (\%) under different modality settings using 5-fold cross-validation. Bold values indicate the best performance.}
\resizebox{\textwidth}{!}{
\begin{tabular}{lllll}
\toprule
Modality & Embedding Space & AUC & ACC  & MCC \\
\midrule

\multirow{2}{*}{MRI Only}
&  Euclidean  & 81.06 $\pm$ 3.25 & 72.21 $\pm$ 6.09 & 43.55 $\pm$ 12.84 \\
& Hyperbolic &  \textbf{84.19} $\pm$ 4.77 & \textbf{75.05} $\pm$ 3.44 & \textbf{58.32} $\pm$ 7.38 \\
\midrule

\multirow{2}{*}{Tabular Only}
& Euclidean  & 85.59 $\pm$ 7.43 & 83.69 $\pm$ 6.47 & 61.02 $\pm$ 13.81 \\
& Hyperbolic & \textbf{86.22} $\pm$ 7.52 & \textbf{85.01} $\pm$ 5.49 & \textbf{62.46} $\pm$ 12.00 \\
\midrule

\multirow{2}{*}{Tabular+MRI}
& Euclidean  & 81.94 $\pm$ 7.07 & 77.25 $\pm$ 3.48 & 46.27 $\pm$ 9.06 \\
& Hyperbolic  & \textbf{90.09} $\pm$ 1.58 & \textbf{83.88} $\pm$ 3.15 & \textbf{67.58} $\pm$ 6.45 \\

\bottomrule
\end{tabular}
}
\label{tab:oasis3-ablation_hyperbolic}
\end{table*}

\noindent\textbf{Effect of hyperbolic representation.} As shown in Table~\ref{tab:oasis3-ablation_hyperbolic}, hyperbolic representations consistently outperform Euclidean representations across all modality settings, but the gain varies across modalities. The tabular-only improvement is modest, with AUC increasing from 85.59\% to 86.22\% and MCC from 61.02\% to 62.46\%, suggesting limited additional gain over the Euclidean baseline because the clinical variables are already highly discriminative. In contrast, the MRI-only setting shows a larger improvement, especially in MCC (43.55\% to 58.32\%), indicating that hyperbolic modelling supports distributed multi-scale anatomical representations. The largest gain appears in the tabular and MRI setting, where AUC increases from 81.94\% to 90.09\% and MCC from 46.27\% to 67.58\%. These results indicate that modelling tabular and MRI features in hyperbolic space benefits both single-modality representation learning and multimodal fusion, with the largest gains observed in joint tabular-MRI fusion.

\begin{table*}[ht]

\centering

\small

\setlength{\tabcolsep}{5pt}

\renewcommand{\arraystretch}{1.0}

\caption{Ablation study of ROI-guided multi-scale MRI sampling. Whole-brain MRI is used as the baseline, and variants compare ROI-guided and random sampling under single- and multi-scale settings. Results are mean $\pm$ standard deviation (\%) from 5-fold cross-validation. Bold indicates the best performance.}

\resizebox{\textwidth}{!}{

\begin{tabular}{llllll}

\toprule

Method & Sampling Strategy & Patch Extraction & AUC & ACC & MCC \\

\midrule

Whole-brain & Whole-brain & Whole-volume 

& 76.66 $\pm$ 13.30 

& 66.29 $\pm$ 18.62 

& 49.85 $\pm$ 25.32 \\

\midrule

\multirow{7}{*}{Variants}

& ROI-guided & Single-scale ($64^3$)

& 89.27 $\pm$ 1.46

& 82.22 $\pm$ 1.43

& 63.77 $\pm$ 2.86 \\

& ROI-guided & Single-scale ($32^3$)

& 89.50 $\pm$ 2.49

& 81.84 $\pm$ 3.34

& 63.04 $\pm$ 6.22 \\

& ROI-guided & Single-scale ($16^3$)

& 89.43 $\pm$ 1.31

& 81.85 $\pm$ 2.08

& 62.81 $\pm$ 4.38 \\

& Random & Multi-scale ($64^3,32^3,16^3$)

& 89.06 $\pm$ 2.24

& 83.51 $\pm$ 3.21

& 66.27 $\pm$ 6.98 \\

& Random & Single-scale ($64^3$)

& 88.99 $\pm$ 2.59

& 83.33 $\pm$ 2.33

& 65.87 $\pm$ 4.96 \\

& Random & Single-scale ($32^3$)

& 89.18 $\pm$ 2.06

& 81.67 $\pm$ 2.86

& 62.95 $\pm$ 5.56 \\

& Random & Single-scale ($16^3$)

& 89.05 $\pm$ 1.65

& 82.77 $\pm$ 2.71

& 65.07 $\pm$ 5.29 \\

\midrule

Ours & ROI-guided & Multi-scale ($64^3,32^3,16^3$)

& \textbf{90.09} $\pm$ 1.58

& \textbf{83.88} $\pm$ 3.15

& \textbf{67.58} $\pm$ 6.45 \\

\bottomrule

\end{tabular}

}

\label{tab:oasis3-ablation-multi-scale}

\end{table*}

\noindent\textbf{Effect of ROI-guided multi-scale MRI sampling.} Table~\ref{tab:oasis3-ablation-multi-scale} evaluates the impact of patch extraction, sampling strategy, and multi-scale MRI encoding. Compared with whole-brain MRI encoding, patch-based variants substantially improve performance, increasing AUC from 76.66\% to around 89\% and ACC from 66.29\% to over 81\%. This suggests that local patch extraction helps suppress irrelevant whole-volume information and improves MRI representation learning. Among single-scale variants, both ROI-guided and random sampling obtain comparable AUC values, while ROI-guided sampling achieves stronger MCC in most cases, indicating more reliable classification. Multi-scale extraction further improves performance by combining complementary anatomical information from different spatial resolutions. The proposed ROI-guided multi-scale strategy achieves the best results, with 90.09\% AUC, 83.88\% ACC, and 67.58\% MCC, outperforming the random multi-scale variant by 1.03\% AUC, 0.37\% ACC, and 1.31\% MCC. These results confirm the effectiveness of ROI-guided multi-scale MRI sampling for dementia classification.

\section{Conclusion}

In this work, we proposed a geometry-aware multimodal framework(HypTI) for dementia classification. By leveraging the hierarchical structure of clinical assessments and the multi-scale anatomical patterns of neurodegeneration, the proposed framework moves beyond conventional Euclidean multimodal fusion. Experimental results show that this design achieves superior performance over representative unimodal and multimodal baselines. Ablation studies demonstrate that hyperbolic representation learning and fusion improve classification performance. These results indicate that the benefits of multimodal dementia detection arise not only from combining heterogeneous modalities, but also from learning representations in a geometry that better matches the underlying structure of the data. This opens new opportunities for geometry-aware multimodal learning in dementia research and broader medical AI applications. In future work, we will extend the evaluation to external datasets and longitudinal clinical measurements, while further analysing the interpretability of hyperbolic representations, and explore multilayer brain networks modelling\cite{bentley2016}, where hyperbolic representations may capture hierarchical cross-layer connectivity patterns.

%
%
%
%
\bibliographystyle{splncs03_unsrt}
\bibliography{paper-camera-ready}

\end{document}